\documentclass[10pt,twocolumn,letterpaper]{article}

\usepackage[pagenumbers]{cvpr} 

\usepackage{graphicx}
\usepackage{amsmath}
\usepackage{amssymb}
\usepackage{bm}
\usepackage{booktabs}

\usepackage{enumitem}
\setlist{nolistsep}
\usepackage[pagebackref,breaklinks,colorlinks]{hyperref}

\usepackage[capitalize]{cleveref}
\crefname{section}{Sec.}{Secs.}
\Crefname{section}{Section}{Sections}
\Crefname{table}{Table}{Tables}
\crefname{table}{Tab.}{Tabs.}

\def\cvprPaperID{4516} 
\def\confName{CVPR}
\def\confYear{2023}

\begin{document}

\title{Multiscale Tensor Decomposition and Rendering Equation Encoding \\for View Synthesis}

\author{Kang Han\\
James Cook University\\
{\tt\small kang.han@my.jcu.edu.au}
\and
Wei Xiang\thanks{Corresponding author.}\\
La Trobe University\\
{\tt\small w.xiang@latrobe.edu.au}
}
\maketitle

\begin{abstract}
Rendering novel views from captured multi-view images has made considerable progress since the emergence of the neural radiance field. This paper aims to further advance the quality of view synthesis by proposing a novel approach dubbed the neural radiance feature field (NRFF). We first propose a multiscale tensor decomposition scheme to organize learnable features so as to represent scenes from coarse to fine scales. We demonstrate many benefits of the proposed multiscale representation, including more accurate scene shape and appearance reconstruction, and faster convergence compared with the single-scale representation. Instead of encoding view directions to model view-dependent effects, we further propose to encode the rendering equation in the feature space by employing the anisotropic spherical Gaussian mixture predicted from the proposed multiscale representation. The proposed NRFF improves state-of-the-art rendering results by over 1 dB in PSNR on both the NeRF and NSVF synthetic datasets. A significant improvement has also been observed on the real-world Tanks \& Temples dataset. Code can be found at \url{https://github.com/imkanghan/nrff}.
\end{abstract}

\section{Introduction}
View synthesis aims to synthesize unrecorded views from multiple captured views using computer vision techniques. A great deal of effort has been made to solve this problem in the past few decades \cite{tewari2022advances}. The recently proposed neural radiance field (NeRF) \cite{mildenhall2020nerf} made a breakthrough in this area by modeling a scene via a multilayer perceptron (MLP). The NeRF achieves an impressive photo-realistic view synthesis quality with 6 degrees of freedom for the first time. The NeRF also represents a scene in a very compact form. That is, only a small number of parameters in the MLP, whose size is even smaller than the captured images. However, this advantage in model size comes at the expense of extensive computations. Numerous evaluations of the MLP are required to render a single pixel, incurring a challenge for both training and testing.

Representing a scene via learnable features is shown to be an effective alternative approach for photo-realistic view synthesis \cite{fridovich2022plenoxels,chen2022tensorf,mueller2022instant,sun2022direct}. Several data structures are employed to efficiently organize learnable features to achieve compact representations. Multiresolution hash encoding (MHE) \cite{mueller2022instant} and tensor decomposition in TensoRF \cite{chen2022tensorf} are two typical works in this direction. MHE organizes learnable features in multiresolution hash tables. As each hash table corresponds to a distinct grid resolution, a point is thus indexed into different positions of the hash tables to mitigate the negative effects of hash collisions. However, this structure breaks the local coherence in nature scenes, even though the spatial hash function in MHE preserves the coherence to some extent. By comparison, TensoRF decomposes a 3D tensor into 2D plane and 1D line tensors, where the local coherence is largely preserved. However, TensoRF's decomposition is performed only in a single scale, whereas multiscale methods are much more desirable for wide-ranging computer vision tasks \cite{takikawa2021neural,aliev2020neural,lin2017feature,sun2019deep,liu2021swin}. We thus propose a multiscale tensor decomposition (MTD) method to represent scenes from coarse to fine scales. We show that the proposed MTD method is able to reconstruct more accurate scene shapes and appearances, and also converges faster than the single-scale TensoRF. As a result, the proposed MTD method achieves better view synthesis quality than TensoRF, even with fewer learnable features.

View direction encoding is the key to the success of neural rendering in modeling complex view-dependent effects. Frequency (or position encoding) \cite{mildenhall2020nerf} and spherical harmonics \cite{verbin2022ref} are the two mostly used view direction encoding methods. The encoded feature vector of a view direction is then fed to an MLP to predict a view-dependent color. This approach models the 5D light field function (3D spatial position with 2D view direction) \cite{levoy1996light}. In computer graphics, the light field is usually modeled by the rendering equation \cite{kajiya1986rendering}, where the outgoing radiance is the interaction result of the incoming light at a point with a specific material. An accurate solution to the rendering equation involves Monte Carlo sampling and integration, which is computationally expensive, especially for the scenario of inverse rendering \cite{hasselgren2022shape}. In this paper, we propose to encode the rendering equation in the feature space in lieu of the color space using the predicted anisotropic spherical Gaussian mixture. In this way, the following MLP is aware of the rendering equation so as to better model complex view-dependent effects. As we use both neural and learnable feature representations as well as the rendering equation encoding (REE) in the feature space, we dub the proposed method the neural radiance feature field (NRFF). In summary, we make the following contributions:
\begin{itemize}
  \item We propose a novel multiscale tensor decomposition scheme to represent scenes from coarse to fine scales, enabling better rendering quality and faster convergence with fewer learnable features;
  \item In lieu of direct encoding of view directions, we propose to encode the rendering equation in the feature space to facilitate the modeling of view-dependent effects.
\end{itemize}

\section{Related work}
We divide view synthesis methods into neural and learnable feature representations depending on whether extra learnable parameters are used to represent a scene in addition to weights and biases in neural networks.

\subsection{Neural representations}
Neural representations mean representing a scene by neural networks, typically MLPs \cite{mildenhall2020nerf} or transformers \cite{suhail2022light}. Mildenhall \etal \cite{mildenhall2020nerf} first proposed this idea for view synthesis in the NeRF and achieved photo-realistic view synthesis results. The MLP in the NeRF is optimized to predict the volume density and the view-dependent appearance of a 3D spatial point observed from a given 2D view direction. Each component in this 5D input is encoded by a set of functions, \eg, sine and cosine, with varying periods before being fed to the MLP. Such position or frequency encoding is one of the key factors to NeRF's success. The input encoding has been further explored in \cite{tancik2020fourier} by a neural tangent kernel and extended in mip-NeRF \cite{barron2021mip} to achieve anti-aliasing view synthesis. Neural representations have the advantage of representing a scene in a very compact form. MLPs are also used to predict the light source visibility of a point to enable relighting \cite{srinivasan2021nerv,zhang2021nerfactor}. However, these methods are computationally expensive because numerous evaluations of the networks are needed to render a single pixel.

Encoding view directions is important for neural representations to achieve photo-realistic view synthesis. Except for the aforementioned position encoding, spherical harmonics are also used to encode view directions with various frequency components \cite{yu2021plenoctrees,verbin2022ref}. This approach composed of view direction encoding and the following MLP modeling is the dominant solution in the current neural rendering approaches. Such view direction encoding methods provide view direction information in various frequencies but neglect the rich information contained in the well-known rendering equation \cite{kajiya1986rendering}. In this paper, instead of encoding view directions, we propose to encode the rendering equation to facilitate the learning of complex view-dependent effects for the following MLP.

\subsection{Learnable feature representations}
Learnable features are parameters that are also optimized by gradient descent in addition to weights and biases in neural networks. Learnable features are usually organized by the data structures of grids, sparse grids, trees, and hash tables. For a given input, interpolation is performed to obtain the corresponding features. The interpolated features can be directly interpreted as some properties, e.g., densities or colors, or optionally fed into neural networks to predict the designed outputs. Compared with pure neural representations, learnable feature representations are computationally efficient at the expense of memory footprint. As the features are also optimized for the considered scene, the task of inferring scene properties for the subsequent MLP is much easier in comparison with predicting from input coordinate encoding. As a result, with learnable feature representations, small MLPs are able to achieve a competitive rendering quality similar to pure neural representations.

Efficient data structures to arrange learnable features are crucial in terms of both computational cost and memory consumption. The 3D dense grid is a significant waste of memory because most of the voxels are empty. Its number of
parameters increases by $\mathcal{O}(N^3)$. Thus, the 3D dense grid is only practical at low resolution, e.g., $N=160$ in \cite{sun2022direct}, limiting its rendering quality. The Octree \cite{liu2020neural,yu2021plenoctrees} and sparse 3D grid \cite{fridovich2022plenoxels} are also employed but data structures need to be updated progressively. Because scene geometry only emerges during training. The recently proposed MHE \cite{mueller2022instant} is a very compact learnable feature representation but hash collision and the break of spatial coherence limit its rendering quality. Concurrent tensor decomposition in TensoRF \cite{chen2022tensorf} preserves spatial coherence but is only performed at a single scale. The benefits of multiscale schemes \cite{lin2017feature,sun2019deep,liu2021swin} studied in the literature inspire us to propose the MTD scheme to represent scenes at varying scales.

\section{Method}
The proposed NRFF obtains the view-dependent color of a point through two main steps. For a point $\mathbf{x}=(x,y,z)$ sampled from a cast ray $\mathbf{r}(t)=\mathbf{o}+t\mathbf{d}$, where $\mathbf{o}$ and $\mathbf{d}$ are the camera center and view direction, respectively, we first compute its feature vector from the proposed multiscale representation. The feature vector is fed into a spatial MLP to predict light parameters used to encode the rendering equation. Next, we apply the proposed REE and then use a directional MLP to predict the final color.

\subsection{Multiscale tensor decomposition}\label{sec:mtd}
\begin{figure}[t]
  \centering
   \includegraphics[width=0.7\linewidth]{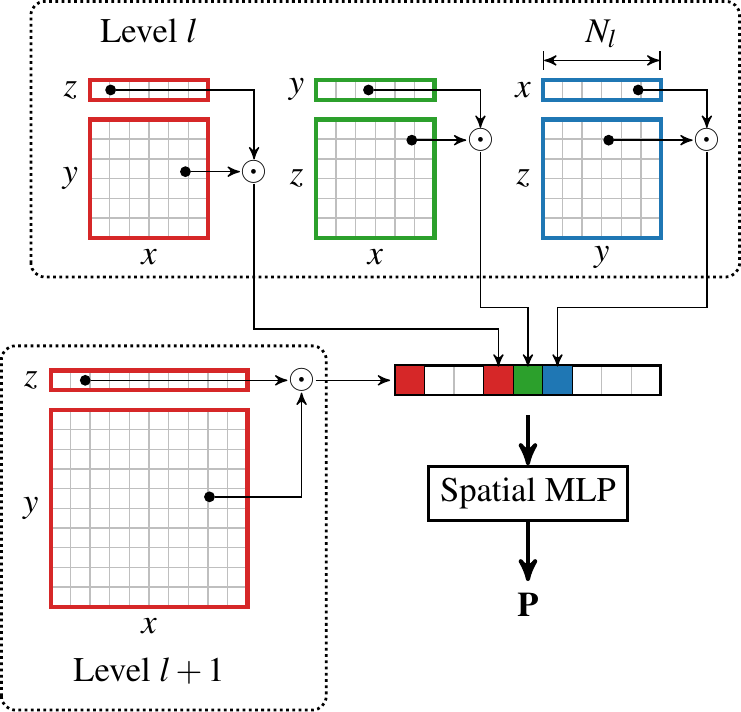}
   \caption{Multiscale tensor decomposition representation. At each level, a 3D tensor representation is decomposed to three sets of plane feature maps and line feature vectors. The resolution of decomposed tensors increases with the level, enabling scene representation at different scales. The concatenated feature vectors from all levels are used to predict parameters $\mathbf{P}$ by a spatial MLP.}
   \label{fig:mtd}
\end{figure}

We propose a multiscale tensor decomposition (MTD) scheme to represent a scene from coarse to fine scales. Similar multiscale ideas have been used in 3D shape representation \cite{takikawa2021neural}, coarse-to-fine point rasterization \cite{aliev2020neural} and many other computer vision works in the literature \cite{lin2017feature, sun2019deep,liu2021swin}. We start with a base resolution of $N_{\text{min}}$ and progressively increase the level resolution to the maximum resolution of $N_{\text{max}}$ by a factor $b$, in line with the strategy in MHE \cite{mueller2022instant}:
\begin{equation}
  N_l = \lfloor N_{\text{min}} b^l \rfloor
\end{equation}
\begin{equation}
  b = \exp\left(\frac{\ln N_{\text{max}} - \ln N_{\text{min}}}{L-1} \right)
\end{equation}
where $N_l$ is the resolution at level $l$ and $L$ is the number of multiscale levels. Feature vectors of point $\mathbf{x}$ are obtained from the proposed MTD independently at different levels. As shown in \cref{fig:mtd}, we use the tensor decomposition mechanism \cite{chen2022tensorf} that decomposes a 3D tensor representation into three plane feature maps and three line feature vectors. We apply linear interpolation (bilinear interpolation for 2D) to the plane feature map $\mathbf{F}_{xy}^l$ and the feature vector $\mathbf{F}_{z}^l$ using the corresponding decomposed coordinates $\mathbf{x}_{xy}, \mathbf{x}_z$ to obtain the following two feature vectors:
\begin{equation}
  \begin{split}
    \mathbf{f}_{xy}^l &= \text{Interp2D}(\mathbf{F}_{xy}^l, \mathbf{x}_{xy}) \\
    \mathbf{f}_{z}^l  &= \text{Interp1D}(\mathbf{F}_{z}^l, \mathbf{x}_{z}). \\
  \end{split}
\end{equation}
The output feature vector at level $l$ is obtained as follows:
\begin{equation}
  \mathbf{f}^l_{xy,z} = \mathbf{f}_{xy}^l\odot\mathbf{f}_{z}^l
  \label{eq:spatial-feature}
\end{equation}
where $\odot$ denotes the element-wise multiplication. Feature vectors from other levels are obtained similarly. The output feature vectors $\left[..., \mathbf{f}^l_{xy,z}, \mathbf{f}^l_{xz,y}, \mathbf{f}^l_{yz,x}, \mathbf{f}^{l+1}_{xy,z}, ...\right]$ from all levels are concatenated and then fed into a spatial MLP to predict parameters $\mathbf{P}$, which will be detailed in \cref{sec:nrff}.

The proposed multiscale scheme brings about three main benefits compared with the single-scale tensor decomposition in TensoRF \cite{chen2022tensorf}. First, it enables better exploration of the local smoothness of nature scenes at varying scales. Coarse-scale representations are inherently smooth, while fine-scale representations provide rich local details. It should be noted that the goal of the multiscale scheme here is different from that of MHE \cite{mueller2022instant}. MHE uses multiresolution mainly for mitigating the negative effects of hash collisions as points are indexed to different positions in the hash tables at varying resolutions. Second, the number of feature channels at each scale could be significantly smaller than that in the single-scale representation, enabling high-resolution representations to explore richer details. For example, a multiscale representation with 16 levels, a maximum resolution of 512, and 4 feature channels has 8.5M parameters, which are fewer than 13M parameters in a single-scale TensoRF with a resolution of 300 and 48 feature channels. In \cref{sec:ablation}, we show that even with fewer parameters, the proposed MTD method outperforms the single-scale TensoRF in terms of rendering quality. Third, scene geometry appears fast in coarse-scale representations, leading to faster convergence than the single-scale representation.

\subsection{Rendering equation encoding}\label{sec:nrff}

A light field can be defined as the radiance at a point in a given direction \cite{levoy1996light}. It is thus represented by a 5D function $L(\mathbf{x}, \bm{\omega}_o)$, where $\mathbf{x}\in \mathbb{R}^3$ is the spatial position and $\bm{\omega}_o\in \mathbb{R}^2$ (spherical coordinate) is the outgoing radiance direction. This 5D light field is the result of the interaction of the scene shape, material, and lighting, which is usually modeled by the rendering equation \cite{kajiya1986rendering} consisting of the diffuse and specular components:
\begin{align}
  L(\bm{\omega}_o; \mathbf{x}) &= \mathbf{c}_d + \mathbf{s}\int_{\Omega} L_i(\bm{\omega}_i; \mathbf{x}) \rho_s(\bm{\omega}_i, \bm{\omega}_o; \mathbf{x})(\mathbf{n}\cdot\bm{\omega}_i) \,d\bm{\omega}_i \notag\\
  &= \mathbf{c}_d + \mathbf{s}\int_{\Omega} f(\bm{\omega}_i, \bm{\omega}_o; \mathbf{x}, \mathbf{n})\,d\bm{\omega}_i
  \label{eq:rendering}
\end{align}
where $\mathbf{c}_d$ indicates the diffuse color and $\mathbf{s}$ is the weight of the specular color. Symbol $\cdot$ indicates the dot product in the Cartesian coordinate system. $L_i$ is the incoming radiance from direction $\bm{\omega}_i$, and $\rho_s$ represents the specular component of the spatially-varying bidirectional reflectance distribution function (BRDF). For ease of exposition, we define $f$ as a function describing the outgoing radiance after the ray interaction. The integral is solved over the hemisphere $\Omega$ defined by the normal vector $\mathbf{n}$ at point $\mathbf{x}$. In computer graphics, $L_i, \rho_s, \mathbf{n}$ are usually known functions or parameters that describe scene lighting, material, and shape. An accurate solution to the rendering equation is achieved by computationally intensive Monte Carlo estimation in the color space, \eg, computing the discrete summation by evaluating $L_i, \rho_s$ at sampled $\bm{\omega}_i$ for a given $\bm{\omega}_o$.

In the inverse rendering problem, $L_i, \rho_s, \mathbf{n}$ are unknown functions or parameters. The most popular method in the inverse rendering to solve the equation is to treat it as a function of $\bm{\omega}_o$, and then employ an MLP to directly predict the integral result from the encoded $\bm{\omega}_o$. However, this simplification neglects the rich information described in the rendering equation and gives the MLP a complicated function to learn. Recent studies have also attempted to estimate the unknown properties to achieve relightable view rendering \cite{boss2021neural, boss2021nerd, zhang2021nerfactor, lyu2022neural}. But their rendering quality is inferior to methods \cite{barron2021mip,verbin2022ref,chen2022tensorf,mueller2022instant} that focus only on view rendering with fixed lighting conditions. We instead propose to encode the rendering equation in the feature space and let the MLP predict the integrated color from the resultant encoding. By doing this, the following MLP is aware of the rendering equation, making the learning task much easier.

\begin{figure*}[t]
  \centering
   \includegraphics[width=0.9\linewidth]{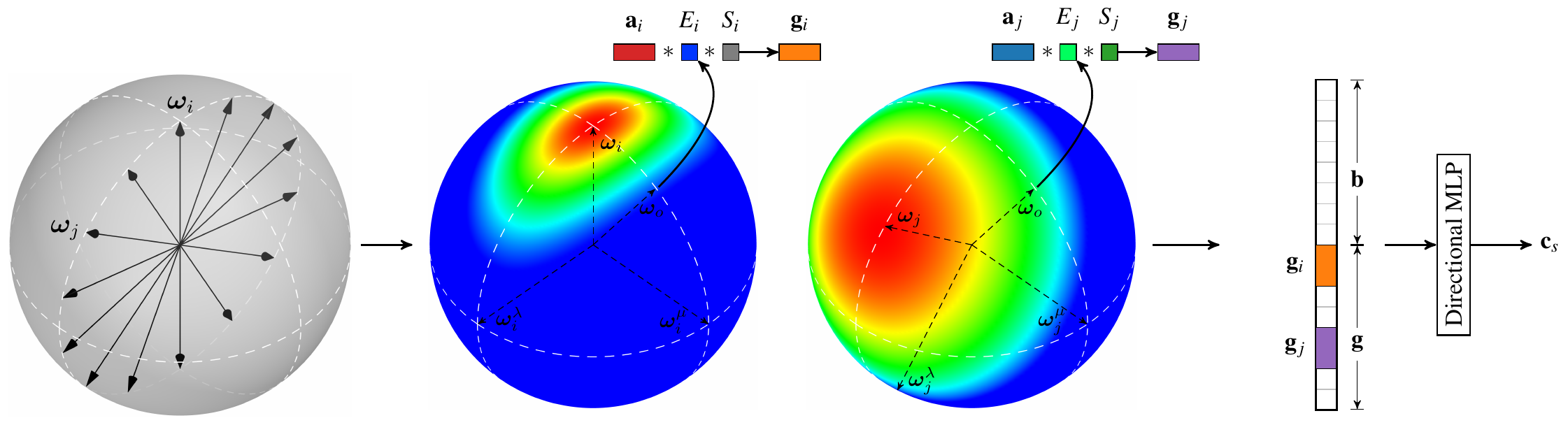}
   \caption{Illustration of the proposed rendering equation encoding. The rendering equation is encoded in the feature space by the learned ASG mixture with predefined orthonormal axes. The axes are defined by a set of radiance directions uniformly sampled on a unit sphere. Here only sampled $\bm{\omega}$ on a plane are shown for better visualization. For a sampled $\bm{\omega}_i$, an ASG function of the reparameterized view direction $\bm{\omega}_o$ is employed to determine the feature response $\mathbf{g}_i$. $E_i$ and $S_i$ are exponential and smooth terms, respectively, and $*$ denotes multiplication. Each ASG function is controlled by learned bandwidths $\lambda$ and $\mu$. The encoded feature vector $\mathbf{g}$ along with a bottleneck feature vector $\mathbf{b}$ depending only on the spatial position, are fed into a directional MLP to predict the specular color $\mathbf{c}_s$.}  
   \label{fig:asg}
\end{figure*}

While encoding the rendering equation in the color space has a clear physical meaning, difficulties in three aspects limit its performance. First, the MLP yields the color parameters in the rendering equation by its final layer. Before the final layer, the MLP does not even know the outgoing radiance direction. This means the MLP does not benefit from the rendering equation as its input does not include information relevant to the rendering equation. Instead, the MLP only learns a spatial function of the position of the input point. Second, using the Monte Carlo integration technique to solve the rendering equation requires many samples to achieve a satisfactory quality in the color space, while extensive sampling is expensive in the inverse rendering problem \cite{zhang2021nerfactor,hasselgren2022shape}. In the feature space, a feature vector consisting of a small number of sampled features could be a comprehensive representation. We show that 128 samples in the feature space are sufficient to render high-quality views. Last, in the color space, approximating the rendering equation by some basis functions (typically spherical Gaussians \cite{xu2013anisotropic,wang2009all} or spherical harmonics \cite{ramamoorthi2006modeling}) leads to a closed-form solution so that sampling over $\bm\omega_i$ can be avoided. However, for the inverse rendering problem, the parameters of the basis functions are unknown and predicted from the MLP. Deriving the final color using the computation (\eg, the product of spherical harmonic coefficients \cite{ramamoorthi2006modeling}) of predicted parameters does not provide much additional useful information for the MLP.

We thus encode the rendering equation in the feature space by viewing $f$ as a function of $\bm{\omega}_o$ for a sampled $\bm{\omega}_i$. In this perspective, we can apply a feature function to each sampled $\bm{\omega}_i$. We choose the basic function based on three considerations: 1) the function shape can be controlled by parameters such that each point can have its independent encoding; 2) the function can model all-frequency information (spherical harmonics are band-limited); 3) the function can be in diverse forms. Thus, we use the anisotropic spherical Gaussian (ASG) \cite{xu2013anisotropic} to encode the rendering equation:
\begin{align}
    \label{eq:asg}
    &\mathbf{c}_s'(\bm{\omega}_o; \mathbf{x}) = \sum_{i=0}^{N-1} G_i(\bm{\omega}_o; \mathbf{x}, [\bm{\omega}_i, \bm{\omega}_i^\lambda, \bm{\omega}_i^\mu], [\lambda_i, \mu_i], \mathbf{a}_i) \\
    &= \sum_{i=0}^{N-1} \mathbf{a}_i S(\bm{\omega}_o; \bm{\omega}_i)\exp\left(-\lambda_i(\bm{\omega}_o\cdot\bm{\omega}_i^\lambda)^2-\mu_i(\bm{\omega}_o\cdot\bm{\omega}_i^\mu)^2\right) \notag
\end{align}
where $\mathbf{c}_s'(\bm{\omega}_o; \mathbf{x})$ is a feature representation of the specular integral in \cref{eq:rendering}; $\mathbf{a}_i$ is a feature vector; $[\bm{\omega}_i, \bm{\omega}_i^\lambda, \bm{\omega}_i^\mu]$ (lobe, tangent and bi-tangent) are predefined orthonormal axes satisfying $\bm{\omega}_i\cdot\bm{\omega}_i^\lambda=\bm{\omega}_i\cdot\bm{\omega}_i^\mu=\bm{\omega}_i^\lambda\cdot\bm{\omega}_i^\mu=0$; $\lambda_i,\mu_i >0$ are the bandwidths for $\bm{\omega}_i^\lambda, \bm{\omega}_i^\mu$ axes, controlling the shape of the ASG function; $S(\bm{\omega}_o; \bm{\omega}_i)=\max(\bm{\omega}_o\cdot\bm{\omega}_i, 0)$ is a smooth term. $G_i$ is thus a function of $\bm{\omega}_o$ defined at the sampled $\bm{\omega}_i$.

A problem of using $\bm\omega_o=-\mathbf{d}$ to encode the rendering equation is that the ASG does not match the behavior of physical specular reflection. According to the law of reflection, the most significant energy from an incoming radiance in direction $\bm\omega_i$ is in the area centered at the reflective direction defined to have the same angle to the surface normal as the incoming radiance, but on the opposite side \cite{haines2021reflection}. However, the energy centers of the ASG functions are in the sampled incoming radiance directions $\bm\omega_i$. We tackle this problem by reparameterizing the view direction to the opposite reflective direction, and treat the reparameterized direction as the outgoing radiance direction $\bm\omega_o$:
\begin{equation}
  \bm{\omega}_o = 2(\mathbf{d}\cdot\mathbf{n})\mathbf{n} - \mathbf{d}.
  \label{eq:reparam}
\end{equation}
After reparameterization, the REE matches the physical specular reflection behavior as $\bm\omega_o$ aligns with $\bm\omega_i$. This reparameterization has also been shown to be able to simplify view interpolation, as studied in \cite{verbin2022ref,wood2000surface}.

As depicted in \cref{fig:asg}, we sample $N=8\times16$ lobes on a unit sphere and determine tangent and bi-tangent axes according to their orthonormal constraint. For a sampled $\bm{\omega}_i=(\theta, \phi)$ in the spherical coordinate system, we define $\bm{\omega}_i^\lambda=(\theta+\pi/2)$ and rotate $\bm{\omega}_i^\lambda$ around $\bm{\omega}_i$ by $\pi/2$ using the quaternion operation to obtain $\bm{\omega}_i^\mu$. Two ASG examples in \cref{fig:asg} show that such ASGs have a strong representation ability to model the rendering equation in the feature space. Simply solving \cref{eq:asg} by computing the sum of encoded feature vectors greatly reduces the channels of the feature representation, limiting its representative ability. Instead, we form a comprehensive feature vector $\mathbf{g}$ by concatenating the encoded feature vectors: 
\begin{equation}
  \mathbf{g} = [\mathbf{g}_0, \mathbf{g}_1, ..., \mathbf{g}_{N-1}].
\end{equation}
Together with a spatial bottleneck feature vector $\mathbf{b}$, we apply a directional MLP to predict the specular color $\mathbf{c}_s$. The required parameters for the ASG encoding are from $\mathbf{P}$, which are predicted by the spatial MLP from the spatial feature vector obtained using \cref{eq:spatial-feature} as aforementioned in \cref{sec:mtd}. In summary, $\mathbf{P}$ include the following parameters: \{$\mathbf{c}_d, \mathbf{s}, \mathbf{n}, \mathbf{b}, \mathbf{a}_i, \lambda_i, \mu_i$\}. Finally, we apply the sigmoid function to obtain the final color:
\begin{equation}
  \mathbf{c} = \text{Sigmoid}(\mathbf{c}_d + \mathbf{s}\odot\mathbf{c}_s).
\end{equation}

We can also interpret the proposed REE as a more advanced view direction encoding method. Our REE has two-fold benefits compared with popular frequency encoding \cite{mildenhall2020nerf} and sphere harmonics \cite{verbin2022ref}. First, every point now has its independent encoding functions controlled by the predicted bandwidths in the ASGs, while the encoding functions are fixed for all points in existing works. Second, a diverse of ASG functions can be produced to achieve much richer encoding compared with a few fixed basis encoding functions in existing methods \cite{mildenhall2020nerf}. In addition, the proposed method can also be seen as an underlying technique to model the surface light field. Our method is more accurate and compact than traditional surface light field methods \cite{wood2000surface}, and provides richer information to the downstream networks compared with recent neural methods that use frequency encoding \cite{oechsle2020learning} or raw view direction \cite{chen2018deep}.

\subsection{Volume rendering}
We use the differentiable volume rendering technique \cite{mildenhall2020nerf} to render a ray according to predicted densities and view-dependent colors. The scene density and appearance fields are modeled separately by two MTD representations. For a point $\mathbf{x}_i$ sampled at depth $t_i$, its density $\sigma_i$ is the result of the softplus activation of the sum of the feature vectors at all levels. The color of the considered point is obtained by the method described in \cref{sec:nrff}. We compute the color composition weights based on densities as follows:
\begin{equation} 
  w_i = \exp\left(-\sum_{j=0}^{i-1}\sigma_j \Delta_j \right)\left(1-\exp(-\sigma_i \Delta_i)\right)
  \label{eq:rendering-weight}
\end{equation}
where $\Delta$ is the sampling interval. We follow the method in TensoRF \cite{chen2022tensorf} that only computes the colors of sampled points whose weights are larger than a predefined threshold. This strategy is effective in reducing the computational cost and makes the appearance representation focus on meaningful points. The rendered pixel color $\hat{\mathbf{c}}$ is a weighted sum of the predicted colors:
\begin{equation}
  \hat{\mathbf{c}} = \sum_{i=0}^{N-1} w_i\mathbf{c}_i.
\end{equation}

\begin{table*}[!ht]
  \centering
  \setlength{\tabcolsep}{0.49pt}
  \begin{tabular}{@{\extracolsep{1.1pt}}lc c c c c c c c c c c c c@{}}
    \toprule
    & &  & &  & \multicolumn{3}{c}{NeRF Synthetic \cite{mildenhall2020nerf}} & \multicolumn{3}{c}{NSVF Synthetic \cite{liu2020neural}}  & \multicolumn{3}{c}{Tanks \& Temples \cite{knapitsch2017tanks}}\\
    \cline{6-8}
    \cline{9-11}
    \cline{12-14}
    & \#Features & \#MLP & Batch size &  Steps & PSNR$\uparrow$ & SSIM$\uparrow$ & LPIPS$\downarrow$ & PSNR$\uparrow$ & SSIM$\uparrow$ & LPIPS$\downarrow$ & PSNR$\uparrow$ & SSIM$\uparrow$ & LPIPS$\downarrow$ \\
    \midrule
    NeRF \cite{mildenhall2020nerf} & N/A & 1,191K &  4096 & 300K & 31.01 & 0.947 & 0.081 & 30.81 & 0.952 & 0.043$^*$ & 25.78 & 0.864 & 0.198$^*$ \\
    Mip-NeRF \cite{barron2021mip}  & N/A & 612K   &  4096 & 1M & 33.09 & 0.961 & 0.043 & - & - & -   & - & - & -\\
    Ref-NeRF \cite{verbin2022ref}  & N/A & 902K   & 16384 & 250K & 33.99 & 0.966 & 0.038 & - & - & - & - & - & -\\
    \midrule
    NSVF \cite{liu2020neural}      & 0.32$\sim$3.2M & 500K & 8192 & 150K & 31.75 & 0.953 & 0.047$^*$ & 35.18 & 0.979 & 0.015$^*$ & 28.48 & 0.901 &  0.155$^*$\\
    DVGO \cite{sun2022direct}      & 49M  & 22K & 8192 & 30K & 31.95 & 0.957 & 0.053 & 35.08 & 0.975 & 0.033  & 28.41 & 0.911 & 0.155 \\
    MHE \cite{mueller2022instant}& 12.6M & 10K &  4096 & 30K & 33.18 & - & - & - & - & -  & - & - & -\\
    TensoRF \cite{chen2022tensorf} & 18.6M &  36K & 4096 & 30K & 33.14 & 0.963 & 0.047 & 36.52 & 0.982 & 0.026  & 28.56 & 0.920 & 0.140 \\
    \midrule
    Ours                           & 12.8M & 549K & 4096 & 30K & 34.65 & 0.975 & 0.034 & 37.76 & 0.986 & 0.019  & 28.87 & 0.927 & 0.127 \\
    Ours                           & 12.8M & 549K & 4096 & 60K & \textbf{35.02} & \textbf{0.977} & \textbf{0.031} & \textbf{38.25} & \textbf{0.988} & \textbf{0.017}  & \textbf{29.05} & \textbf{0.931} & \textbf{0.119} \\
    \bottomrule
  \end{tabular}
  \caption{Objective performance comparison. \# denotes the number of learnable parameters. The LPIPS are evaluated using the VGG network, while $^*$ means results from the Alex network. Our LPIPS results with 60K training steps evaluated by the Alex network on the three datasets are 0.016, 0.007, and 0.092, respectively.}
  \label{tab:res-objective}
\end{table*}

\subsection{Training loss}
The training loss of the proposed method consists of the mean squared error of the rendered pixel value, a regularization term about the predicted surface normals \cite{verbin2022ref}, and a regularization term regarding the density features \cite{chen2022tensorf}. Mathematically, the training loss is written as:
\begin{equation}
  \mathcal{L} = (\hat{\mathbf{c}} - \mathbf{c}_{gt}) + \alpha\frac{1}{N} \sum_{i=0}^{N-1} w_i \max(0, {\mathbf{d}}\cdot{\mathbf{n}}_i)^2 + \beta\frac{1}{M} \sum_{i=0}^{M-1} |\mathbf{F}_\sigma^i|
\end{equation}
where $\mathbf{c}_{gt}$ is the ground truth color, $N$ represents the number of samples in the cast ray, and $M$ is the number of features in the density field representation. The normal regularization term, i.e., the second term in the above equation, penalizes the densities which decrease along the ray. In other words, it encourages concentrated modeling of the scene surface. The third term is density regularization defined as the mean absolute value of all features, which encourages a sparse density field. $\alpha,\beta$ are loss weights to balance the impact of the two regularization terms, and we empirically use $\alpha=0.3$ and $\beta=0.0004$ for all experiments as in \cite{verbin2022ref,chen2022tensorf}.

\section{Experiments}

We implement the proposed method using PyTorch \cite{paszke2019pytorch}. There are a total of 16 levels starting with a base resolution of 16 and growing to a maximum resolution of 512. The number of feature channels is 4 for the appearance field and 2 for the density field. The sizes of the bottleneck $\textbf{b}$ and feature vector $\mathbf{a}_i$ are 128 and 2, respectively. The spatial MLP has 3 layers, while the directional one has 6 layers. All layers contain 256 hidden units and ReLU activation. We optimize the proposed model using the Adam algorithm \cite{kingma2014adam} with a learning rate of 2e-3 for the MTDs and 1e-3 for two MLPs. The learning rates degrade log-linearly to 0.1 times their initial values.

We compare our method with methods based on both neural representations and learnable feature representations. The compared methods based on neural representations include NeRF \cite{mildenhall2020nerf}, Mip-NeRF \cite{barron2021mip}, and Ref-NeRF \cite{verbin2022ref}, while NSVF \cite{liu2020neural}, DVGO \cite{sun2022direct}, MHE \cite{mueller2022instant}, and TensoRF \cite{chen2022tensorf} belong to learnable feature representations. We evaluate the rendering quality of these methods using the PSNR, SSIM \cite{wang2004image}, and LPIPS \cite{zhang2018unreasonable}. Two synthetic datasets, namely the NeRF synthetic \cite{mildenhall2020nerf} and NSVF synthetic \cite{liu2020neural} datasets, and one real-world Tanks \& Temples dataset \cite{knapitsch2017tanks} are used for evaluation. Model details including the number of parameters of learnable features and MLPs, batch size, and training steps are also presented for comparison.

\subsection{Objective results}
The proposed method significantly outperforms existing state-of-the-art view synthesis approaches as shown in \cref{tab:res-objective}. Over 1 dB improvement in PSNR has been observed on both the NeRF and NSVF synthetic datasets. Pure MLP-based methods are compact in representing a scene but are computationally expensive. Besides, they also require a large number of training steps to converge. For example, Ref-NeRF \cite{verbin2022ref} takes 250K steps to converge when using a large batch size of 16384. Thanks to the proposed MTD and encoding the rendering equation in the feature space, we are able to use 12.8M learnable features, which is similar to that in MHE \cite{mueller2022instant} and fewer than those in DVGO \cite{sun2022direct} and TensoRF \cite{chen2022tensorf}, to achieve significantly better rendering quality than those compared methods. The proposed NRFF also outperforms the compared methods on the Tanks \& Temples dataset, demonstrating the efficacy of our method in representing real-world scenes.

\begin{figure*}[ht]
  \centering
   \includegraphics[width=\textwidth]{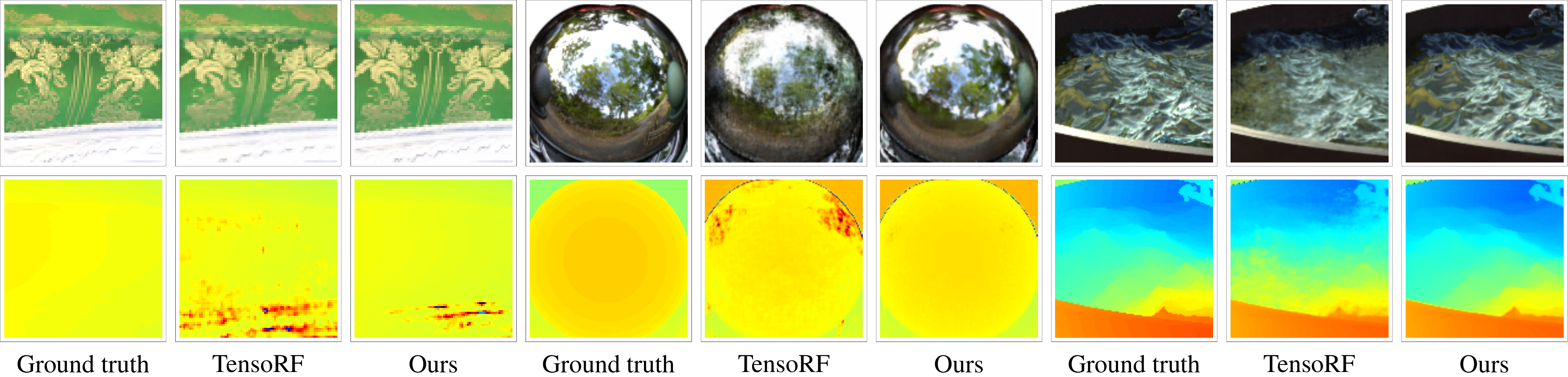}
   \caption{Subjective comparison of rendered views. The first row shows rendered novel views and the second row shows their corresponding depth maps. Our method recovers more accurate texture, specular surface, and geometry than TensoRF \cite{chen2022tensorf}. The scenes from left to right are \textit{chair}, \textit{materials}, and \textit{ship} from the NeRF synthetic dataset \cite{mildenhall2020nerf}.}
   \label{fig:subjective}
\end{figure*}

\subsection{Subjective results}

\begin{figure}[t]
  \centering
   \includegraphics[width=\linewidth]{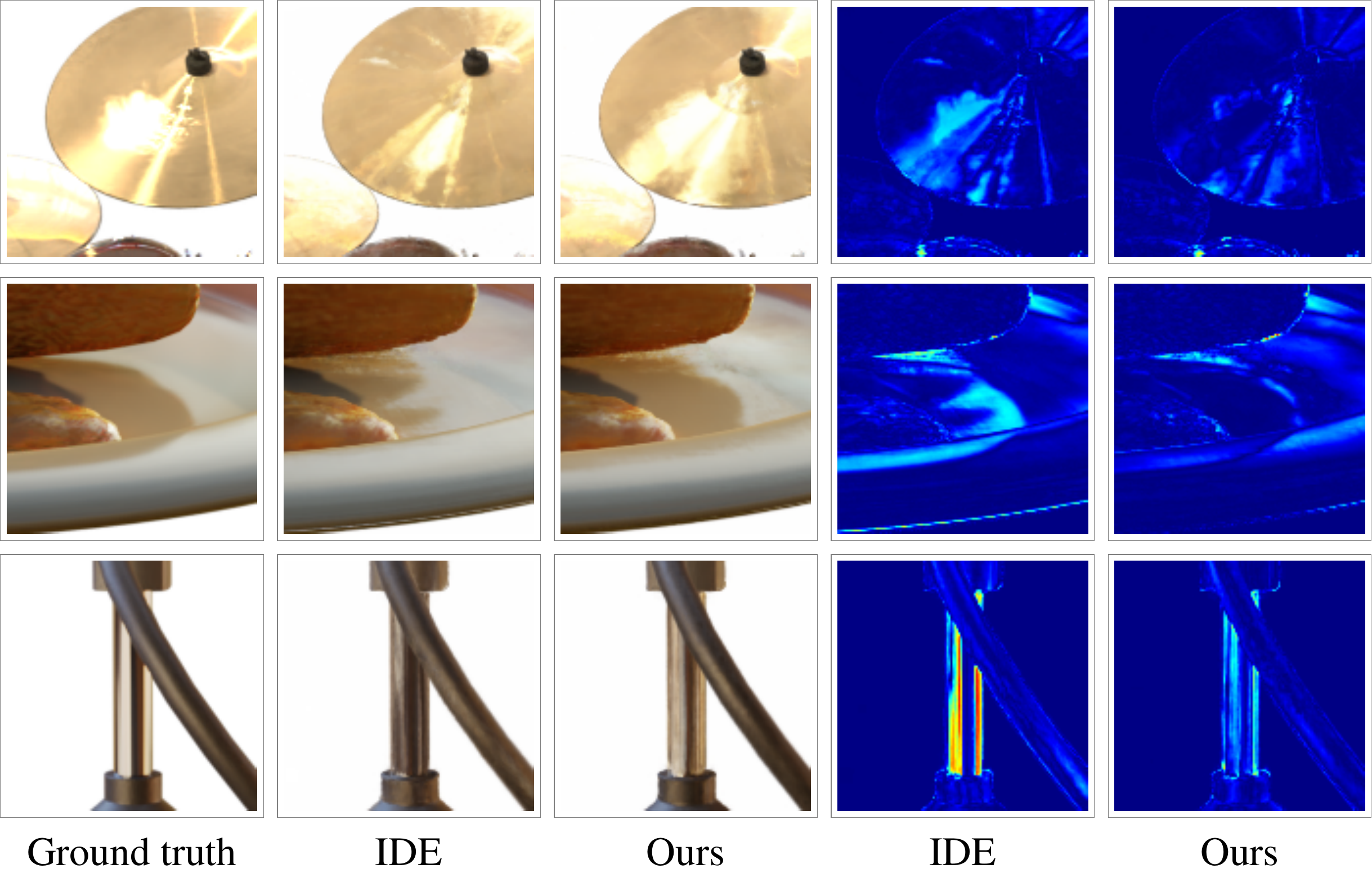}
   \caption{Visual comparison of two methods in modeling view-dependent effects. For fair comparison, all components are fixed except for the methods used to model view-dependent effects. From top to bottom: \textit{drums}, \textit{hotdog}, and \textit{mic} from \cite{mildenhall2020nerf}.}
   \label{fig:ide-vs-ree}
\end{figure}

\begin{figure}[t]
  \centering
   \includegraphics[width=\linewidth]{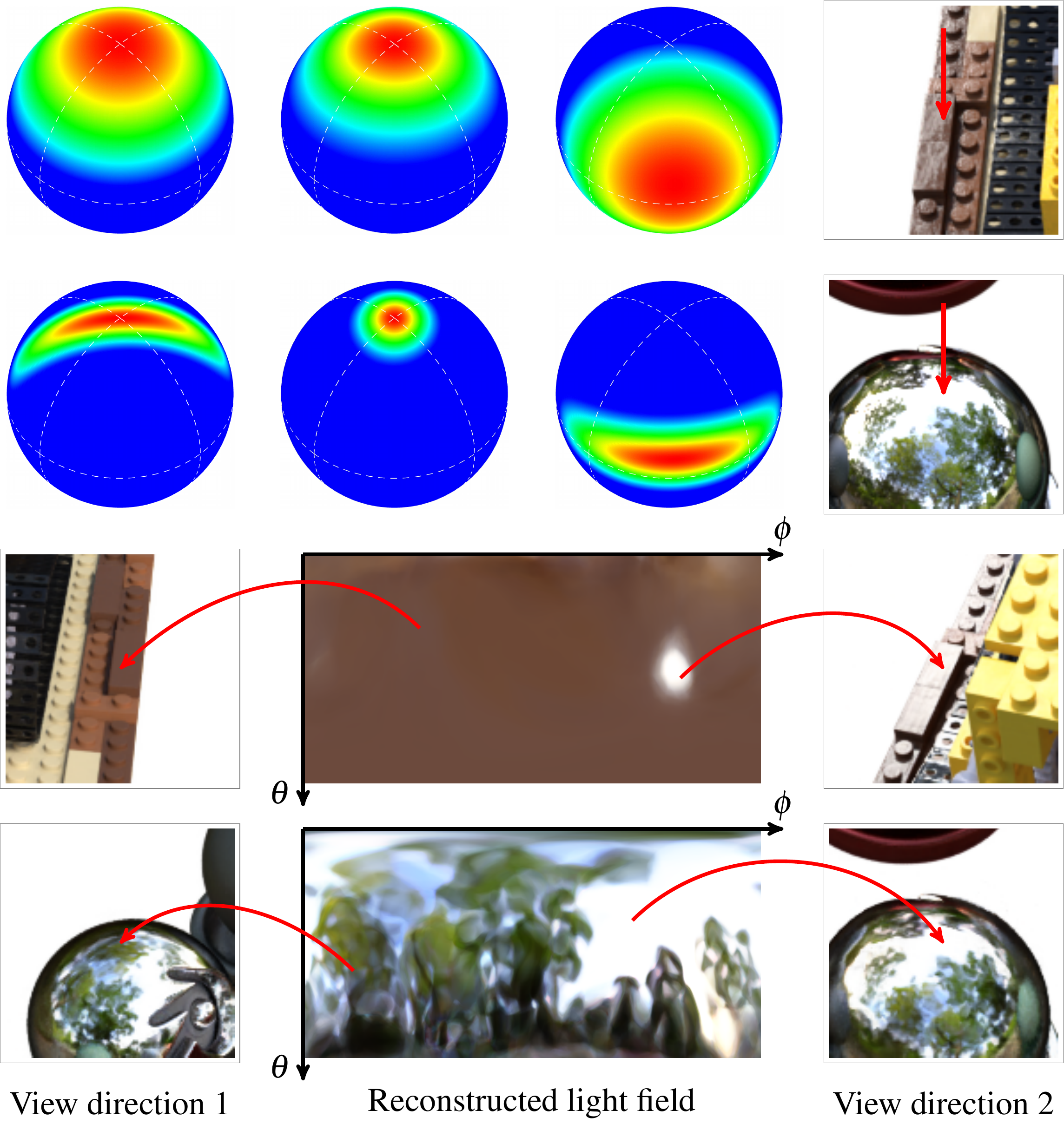}
   \caption{Visualization of the learned ASG functions and reconstructed light fields on the scenes of \textit{lego} and \textit{materials}. The first and second rows show the learned ASG functions used to encode the rendering equation at two points, which are on the rays cast from the pixels' position indicated by the red arrows in the rightmost image patches in the first two rows. The two points have their independent and diverse ASG functions. Our model produces more complex ASG functions on the specular surface (second row) to model the complex reflections. The reconstructed light fields and rendered images at different view directions imply successful modeling of complex view-dependent effects.}
   \label{fig:asg-vis}
\end{figure}

Subjective comparisons are presented in \cref{fig:subjective} to show that our method is able to recover accurate texture, specular surface, and geometry. For fair comparison, we use the results from TensoRF with decreased features (as detailed in the ablation study in \cref{sec:ablation}) such that the model has a similar number of parameters in the learnable features and the MLP as ours. The comparison on scene $chair$ in \cref{fig:subjective} shows that our method synthesizes sharper texture than TensoRF. This advantage stems from the high-resolution representation in our method, which provides rich local details for view synthesis. The rendered balls in the scene \textit{materials} demonstrate the superiority of our REE method in modeling the specular surface compared with the position encoding of view directions employed in TensoRF \cite{chen2022tensorf}. Our multiscale representation also enables more accurate geometry reconstruction as shown in the depth map of the scene \textit{ship}, resulting in more realistic view synthesis of the water surface. The visual comparison in \cref{fig:ide-vs-ree} demonstrates that the proposed REE produces better reflection and illumination effects than the integrated directional encoding (IDE) proposed in Ref-NeRF \cite{verbin2022ref}. Lastly, it is observed from \cref{fig:asg-vis} that our model yields a diverse of ASG functions to encode the rendering equation and reconstructs accurate light fields of scenes.

\subsection{Ablation study}\label{sec:ablation}

We investigate the effectiveness of the proposed modules in \cref{tab:ablation}. We start with the single-scale TensoRF \cite{chen2022tensorf}. All reported results in this table are from models trained by 60K steps. Simply increasing the MLP's size in TensoRF to 10 layers greatly improves the rendering quality but also increases the training and testing times. This quality improvement highlights that both the learnable features and MLP are important for improving the rendering quality. When we decrease the number of learnable features in TensoRF to the same level as in our model, there is a small performance degradation (around 0.1 dB). As our encoding method produces a comparable larger encoding vector, for fair comparison, our models use MLPs with 9 layers to make the MLPs' parameters roughly consistent or fewer than that in TensoRF using 10 layers. Our MTD using the position encoding (PE) to encode view directions achieves better rendering quality than the single-scale TensoRF, even with fewer learnable features and a smaller MLP. 

\begin{table}[t]
  \centering
  \footnotesize
  \setlength{\tabcolsep}{1.0pt}
  \begin{tabular}{@{}lcccccccc@{}}
    \toprule
     & \#Feat. & \#MLP & \#L & Train & Test & PSNR$\uparrow$ & SSIM$\uparrow$ & LPIPS$\downarrow$ \\
    \midrule
    \footnotesize{TensoRF} \cite{chen2022tensorf} & 18.6M & 36K & 4 & 1.34h & 0.99s & 33.43 & 0.964 & 0.045\\
    \footnotesize{TensoRF, inc. MLP} & 18.6M & 568K & 10 & 2.02h & 1.71s & 34.10 & 0.970 & 0.038  \\
    \footnotesize{TensoRF, dec. feat.} & 13.4M & 557K & 10 & 2.19h & 1.72s & 33.99 & 0.969 & 0.039 \\
    \midrule
    \footnotesize{Ours, MTD, PE} & 12.8M & 515K & 9 & 3.32h & 1.92s & 34.58 & 0.975 & 0.034  \\
    \footnotesize{Ours, MTD, IDE} & 12.8M & 532K & 9 & 3.52h & 2.04s & 34.61 & 0.974 & 0.034 \\
    \footnotesize{Ours, MTD, color} & 12.8M & 545K & 9 & 3.36h & 2.06s & 33.53 & 0.965 & 0.043 \\
    \footnotesize{Ours, full} & 12.8M & 549K & 9 & 3.39h & 2.09s & \textbf{35.02} & \textbf{0.977} & \textbf{0.031} \\
    \bottomrule
  \end{tabular}
  \caption{Ablation study on the NeRF synthetic dataset \cite{mildenhall2020nerf}. \#L indicates the number of MLP layers. Training and testing times are averaged over all scenes and frames, respectively.}
  \label{tab:ablation}
\end{table}

Further quality improvement is observed when using our MTD in conjunction with the proposed REE method (i.e., ours, full). As can be observed from \cref{tab:ablation} that our full model improves the PSNR from 34.58 dB (ours, MTD, PE) to 35.02 dB, yielding the state-of-the-art rendering quality. We also experiment on the IDE \cite{verbin2022ref} using our MTD as the input coordinate encoding instead of integrated positional encoding in mip-NeRF \cite{barron2021mip}. We do not observe a significant performance improvement when using the IDE method. The model (ours, MTD, color) using the same form of the REE but in the color space performs poorly. This verifies the drawbacks of encoding the rendering equation in the color space, as discussed in \cref{sec:nrff}.

\begin{figure}[t]
  \centering
   \includegraphics[width=\linewidth]{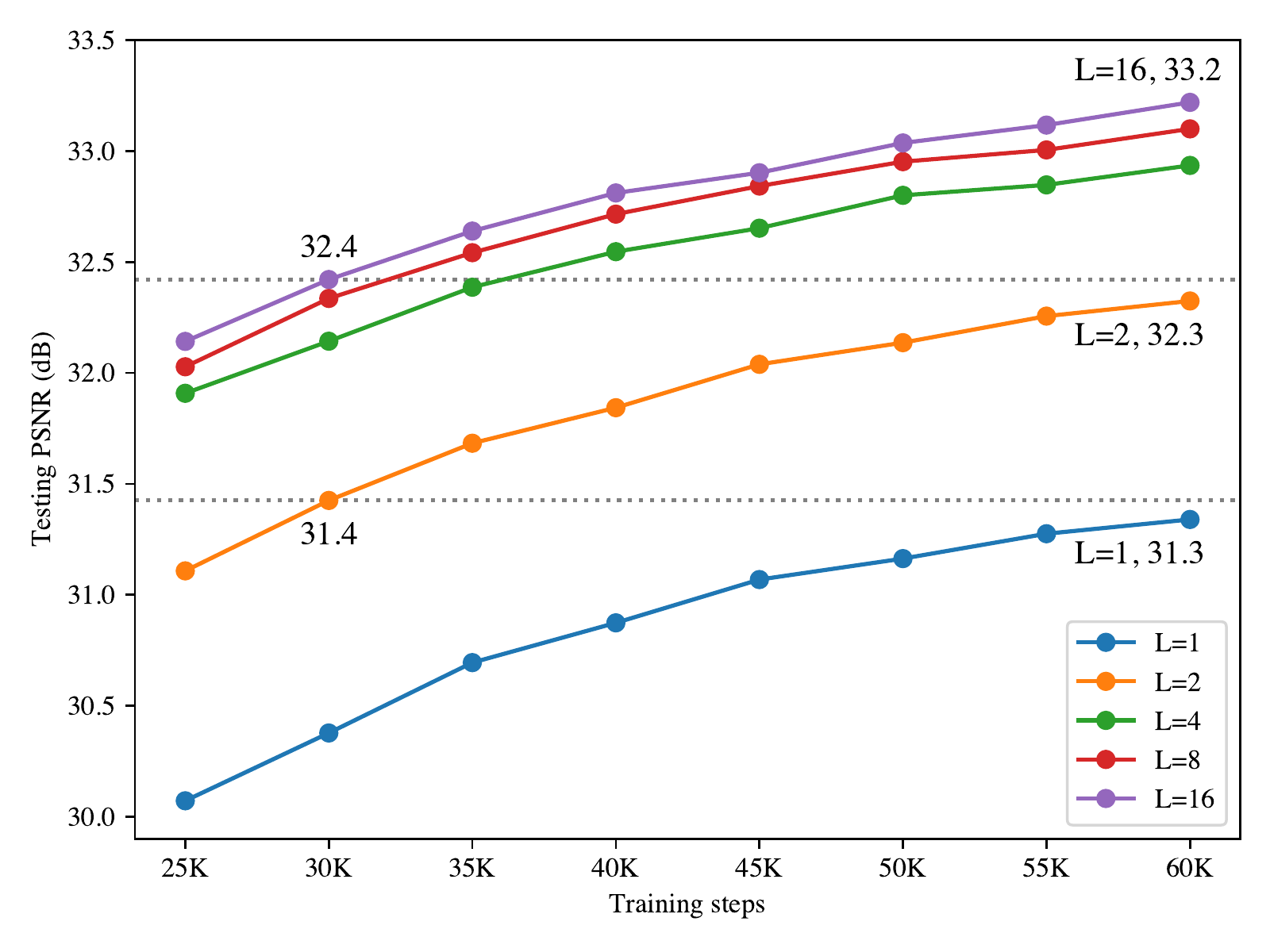}
   \caption{Performance comparison over training steps for varying the number of scale levels on scene \textit{ship} from \cite{mildenhall2020nerf}. $L$ indicates the number of levels. All models with different levels have roughly the same number of learnable features. Models with more levels not only converge faster but yield better final PSNRs.}
   \label{fig:ablation-levels}
\end{figure}

A detailed performance evaluation over training steps for varying the number of scale levels in \cref{fig:ablation-levels} demonstrates the benefits of the proposed MTD scheme. For each setup, we adjust the number of feature channels and the maximum resolution to keep roughly the same number of parameters (12.8M) in the learnable features. As shown in \cref{fig:ablation-levels}, the model with two levels trained by 30K steps already surpasses that with one level trained by 60K steps, suggesting faster convergence speed of the multiscale representation than its single-scale counterpart. 1 dB improvement of the final PSNR is observed (from 31.3 dB with $L=1$ to 32.3 dB with $L=2$) when we have the two-level representation. Nearly 1 dB additional performance gain (from 32.3 dB with $L=2$ to 33.2 dB with $L=16$) becomes observable when increasing the number of levels to 16. 

\subsection{Limitations}
Our method use a comparable large MLP than popular methods \cite{chen2022tensorf,mueller2022instant} with learnable features. Representations with more scale levels introduce extra computations for interpolation weights compared with single-scale representation. On the NeRF synthetic dataset, training takes 3$\sim$4 hours for each scene on one Nvidia Tesla V100 with 32 GB memory, and rendering an image of resolution 800$\times$800 requires 2$\sim$3 seconds. When using MLPs with similar size, the training time of our method is longer than the single-scale TensoRF but our rendering time is only slightly increased as presented in \cref{tab:ablation}. The speed of the proposed method is slower than fast methods \cite{mueller2022instant,sun2022direct}, but faster than pure MLP methods \cite{verbin2022ref,barron2021mip,mildenhall2020nerf}. We believe thorough optimization could overcome this limitation to some extent, considering that the hash encoding in \cite{mueller2022instant} is fast thanks to the highly efficient implementation even with trilinear interpolation. Besides, in the testing stage, the plane feature maps could also be loaded to GPU texture memory to leverage hardware accelerated bilinear interpolation to fetch features more efficiently.

\section{Conclusion}
We proposed the novel neural radiance feature field (NRFF) to achieve photo-realistic view synthesis. The proposed multiscale tensor decomposition scheme represents scenes from coarse to fine scales, leading to faster convergence and a better rendering quality than the single-scale tensor decomposition. Our proposed rendering equation encoding in the feature space provides more knowledge about the outgoing radiance to the MLP and overcomes the limitations of encoding the rendering equation in the color space. Extensive experimental results were presented to demonstrate the efficacy of the proposed NRFF on both the synthetic and real-world datasets.

{\small
\bibliographystyle{ieee_fullname}
\bibliography{egbib}

\begin{thebibliography}{10}\itemsep=-1pt

\bibitem{aliev2020neural}
Kara-Ali Aliev, Artem Sevastopolsky, Maria Kolos, Dmitry Ulyanov, and Victor
  Lempitsky.
\newblock Neural point-based graphics.
\newblock In {\em ECCV}, pages 696--712, 2020.

\bibitem{barron2021mip}
Jonathan~T Barron, Ben Mildenhall, Matthew Tancik, Peter Hedman, Ricardo
  Martin-Brualla, and Pratul~P Srinivasan.
\newblock Mip-{NeRF}: A multiscale representation for anti-aliasing neural
  radiance fields.
\newblock In {\em ICCV}, pages 5855--5864, 2021.

\bibitem{boss2021nerd}
Mark Boss, Raphael Braun, Varun Jampani, Jonathan~T Barron, Ce Liu, and Hendrik
  Lensch.
\newblock {NeRD}: Neural reflectance decomposition from image collections.
\newblock In {\em ICCV}, pages 12684--12694, 2021.

\bibitem{boss2021neural}
Mark Boss, Varun Jampani, Raphael Braun, Ce Liu, Jonathan Barron, and Hendrik
  Lensch.
\newblock {Neural-PIL}: Neural pre-integrated lighting for reflectance
  decomposition.
\newblock {\em NIPS}, 34:10691--10704, 2021.

\bibitem{chen2018deep}
Anpei Chen, Minye Wu, Yingliang Zhang, Nianyi Li, Jie Lu, Shenghua Gao, and
  Jingyi Yu.
\newblock Deep surface light fields.
\newblock {\em Proceedings of the ACM on Computer Graphics and Interactive
  Techniques}, 1(1):1--17, 2018.

\bibitem{chen2022tensorf}
Anpei Chen, Zexiang Xu, Andreas Geiger, Jingyi Yu, and Hao Su.
\newblock {TensoRF}: Tensorial radiance fields.
\newblock {\em ECCV}, 2022.

\bibitem{fridovich2022plenoxels}
Sara Fridovich-Keil, Alex Yu, Matthew Tancik, Qinhong Chen, Benjamin Recht, and
  Angjoo Kanazawa.
\newblock Plenoxels: Radiance fields without neural networks.
\newblock In {\em CVPR}, pages 5501--5510, 2022.

\bibitem{haines2021reflection}
Eric Haines.
\newblock Reflection and refraction formulas.
\newblock In {\em Ray Tracing Gems II}, pages 105--108. Springer, 2021.

\bibitem{hasselgren2022shape}
Jon Hasselgren, Nikolai Hofmann, and Jacob Munkberg.
\newblock Shape, light \& material decomposition from images using monte carlo
  rendering and denoising.
\newblock {\em arXiv preprint arXiv:2206.03380}, 2022.

\bibitem{kajiya1986rendering}
James~T Kajiya.
\newblock The rendering equation.
\newblock In {\em Proceedings of the 13th Annual Conference on Computer
  Graphics and Interactive Techniques}, pages 143--150, 1986.

\bibitem{kingma2014adam}
Diederik~P Kingma and Jimmy Ba.
\newblock Adam: A method for stochastic optimization.
\newblock {\em arXiv preprint arXiv:1412.6980}, 2014.

\bibitem{knapitsch2017tanks}
Arno Knapitsch, Jaesik Park, Qian-Yi Zhou, and Vladlen Koltun.
\newblock Tanks and temples: Benchmarking large-scale scene reconstruction.
\newblock {\em ACM Transactions on Graphics (ToG)}, 36(4):1--13, 2017.

\bibitem{levoy1996light}
Marc Levoy and Pat Hanrahan.
\newblock Light field rendering.
\newblock In {\em Proceedings of the 23rd Annual Conference on Computer
  Graphics and Interactive Techniques}, pages 31--42, 1996.

\bibitem{lin2017feature}
Tsung-Yi Lin, Piotr Doll{\'a}r, Ross Girshick, Kaiming He, Bharath Hariharan,
  and Serge Belongie.
\newblock Feature pyramid networks for object detection.
\newblock In {\em CVPR}, pages 2117--2125, 2017.

\bibitem{liu2020neural}
Lingjie Liu, Jiatao Gu, Kyaw Zaw~Lin, Tat-Seng Chua, and Christian Theobalt.
\newblock Neural sparse voxel fields.
\newblock {\em NIPS}, 33:15651--15663, 2020.

\bibitem{liu2021swin}
Ze Liu, Yutong Lin, Yue Cao, Han Hu, Yixuan Wei, Zheng Zhang, Stephen Lin, and
  Baining Guo.
\newblock Swin transformer: Hierarchical vision transformer using shifted
  windows.
\newblock In {\em ICCV}, pages 10012--10022, 2021.

\bibitem{lyu2022neural}
Linjie Lyu, Ayush Tewari, Thomas Leimk{\"u}hler, Marc Habermann, and Christian
  Theobalt.
\newblock Neural radiance transfer fields for relightable novel-view synthesis
  with global illumination.
\newblock In {\em ECCV}, page 153–169, 2022.

\bibitem{mildenhall2020nerf}
Ben Mildenhall, Pratul~P Srinivasan, Matthew Tancik, Jonathan~T Barron, Ravi
  Ramamoorthi, and Ren Ng.
\newblock {NeRF}: Representing scenes as neural radiance fields for view
  synthesis.
\newblock In {\em ECCV}, pages 405--421, 2020.

\bibitem{mueller2022instant}
Thomas M\"uller, Alex Evans, Christoph Schied, and Alexander Keller.
\newblock Instant neural graphics primitives with a multiresolution hash
  encoding.
\newblock {\em ACM Transactions on Graphics (TOG)}, 41(4):102:1--102:15, July
  2022.

\bibitem{oechsle2020learning}
Michael Oechsle, Michael Niemeyer, Christian Reiser, Lars Mescheder, Thilo
  Strauss, and Andreas Geiger.
\newblock Learning implicit surface light fields.
\newblock In {\em 2020 International Conference on 3D Vision (3DV)}, pages
  452--462. IEEE, 2020.

\bibitem{paszke2019pytorch}
Adam Paszke, Sam Gross, Francisco Massa, Adam Lerer, James Bradbury, Gregory
  Chanan, Trevor Killeen, Zeming Lin, Natalia Gimelshein, Luca Antiga, et~al.
\newblock Pytorch: An imperative style, high-performance deep learning library.
\newblock {\em NIPS}, 32, 2019.

\bibitem{ramamoorthi2006modeling}
Ravi Ramamoorthi.
\newblock Modeling illumination variation with spherical harmonics.
\newblock {\em Face Processing: Advanced Modeling Methods}, pages 385--424,
  2006.

\bibitem{srinivasan2021nerv}
Pratul~P Srinivasan, Boyang Deng, Xiuming Zhang, Matthew Tancik, Ben
  Mildenhall, and Jonathan~T Barron.
\newblock Nerv: Neural reflectance and visibility fields for relighting and
  view synthesis.
\newblock In {\em CVPR}, pages 7495--7504, 2021.

\bibitem{suhail2022light}
Mohammed Suhail, Carlos Esteves, Leonid Sigal, and Ameesh Makadia.
\newblock Light field neural rendering.
\newblock In {\em CVPR}, pages 8269--8279, 2022.

\bibitem{sun2022direct}
Cheng Sun, Min Sun, and Hwann-Tzong Chen.
\newblock Direct voxel grid optimization: Super-fast convergence for radiance
  fields reconstruction.
\newblock In {\em CVPR}, pages 5459--5469, 2022.

\bibitem{sun2019deep}
Ke Sun, Bin Xiao, Dong Liu, and Jingdong Wang.
\newblock Deep high-resolution representation learning for human pose
  estimation.
\newblock In {\em CVPR}, pages 5693--5703, 2019.

\bibitem{takikawa2021neural}
Towaki Takikawa, Joey Litalien, Kangxue Yin, Karsten Kreis, Charles Loop, Derek
  Nowrouzezahrai, Alec Jacobson, Morgan McGuire, and Sanja Fidler.
\newblock Neural geometric level of detail: Real-time rendering with implicit
  3d shapes.
\newblock In {\em CVPR}, pages 11358--11367, 2021.

\bibitem{tancik2020fourier}
Matthew Tancik, Pratul Srinivasan, Ben Mildenhall, Sara Fridovich-Keil, Nithin
  Raghavan, Utkarsh Singhal, Ravi Ramamoorthi, Jonathan Barron, and Ren Ng.
\newblock Fourier features let networks learn high frequency functions in low
  dimensional domains.
\newblock In {\em NIPS}, pages 7537--7547, 2020.

\bibitem{tewari2022advances}
Ayush Tewari, Justus Thies, Ben Mildenhall, Pratul Srinivasan, Edgar Tretschk,
  W Yifan, Christoph Lassner, Vincent Sitzmann, Ricardo Martin-Brualla, Stephen
  Lombardi, et~al.
\newblock Advances in neural rendering.
\newblock {\em Computer Graphics Forum}, 41(2):703--735, 2022.

\bibitem{verbin2022ref}
Dor Verbin, Peter Hedman, Ben Mildenhall, Todd Zickler, Jonathan~T Barron, and
  Pratul~P Srinivasan.
\newblock Ref-{NeRF}: Structured view-dependent appearance for neural radiance
  fields.
\newblock In {\em CVPR}, pages 5491--5500, 2022.

\bibitem{wang2009all}
Jiaping Wang, Peiran Ren, Minmin Gong, John Snyder, and Baining Guo.
\newblock All-frequency rendering of dynamic, spatially-varying reflectance.
\newblock In {\em ACM SIGGRAPH Asia}, pages 1--10, 2009.

\bibitem{wang2004image}
Zhou Wang, Alan~C Bovik, Hamid~R Sheikh, and Eero~P Simoncelli.
\newblock Image quality assessment: from error visibility to structural
  similarity.
\newblock {\em IEEE Transactions on Image Processing}, 13(4):600--612, 2004.

\bibitem{wood2000surface}
Daniel~N Wood, Daniel~I Azuma, Ken Aldinger, Brian Curless, Tom Duchamp,
  David~H Salesin, and Werner Stuetzle.
\newblock Surface light fields for {3D} photography.
\newblock In {\em Proceedings of the 27th Nnnual Conference on Computer
  Graphics and Interactive Techniques}, pages 287--296, 2000.

\bibitem{xu2013anisotropic}
Kun Xu, Wei-Lun Sun, Zhao Dong, Dan-Yong Zhao, Run-Dong Wu, and Shi-Min Hu.
\newblock Anisotropic spherical gaussians.
\newblock {\em ACM Transactions on Graphics (TOG)}, 32(6):1--11, 2013.

\bibitem{yu2021plenoctrees}
Alex Yu, Ruilong Li, Matthew Tancik, Hao Li, Ren Ng, and Angjoo Kanazawa.
\newblock {PlenOctrees} for real-time rendering of neural radiance fields.
\newblock In {\em ICCV}, pages 5752--5761, 2021.

\bibitem{zhang2018unreasonable}
Richard Zhang, Phillip Isola, Alexei~A Efros, Eli Shechtman, and Oliver Wang.
\newblock The unreasonable effectiveness of deep features as a perceptual
  metric.
\newblock In {\em CVPR}, pages 586--595, 2018.

\bibitem{zhang2021nerfactor}
Xiuming Zhang, Pratul~P Srinivasan, Boyang Deng, Paul Debevec, William~T
  Freeman, and Jonathan~T Barron.
\newblock {NeRFactor}: Neural factorization of shape and reflectance under an
  unknown illumination.
\newblock {\em ACM Transactions on Graphics (TOG)}, 40(6):1--18, 2021.

\end{thebibliography}
}

\end{document}